# Object-oriented SLAM using Quadrics and Symmetry Properties for Indoor Environments

Ziwei Liao[1], Wei Wang[1,*], Xianyu Qi[1], Xiaoyu Zhang[1], Lin Xue[2], Jianzhen Jiao[2] and Ran Wei[2]

*Abstract*— Aiming at the application environment of indoor mobile robots, this paper proposes a sparse object-level SLAM algorithm based on an RGB-D camera. A quadric representation is used as a landmark to compactly model objects, including their position, orientation, and occupied space. The state-of-art quadric-based SLAM algorithm faces the observability problem caused by the limited perspective under the plane trajectory of the mobile robot. To solve the problem, the proposed algorithm fuses both object detection and point cloud data to estimate the quadric parameters. It finishes the quadric initialization based on a single frame of RGB-D data, which significantly reduces the requirements for perspective changes. As objects are often observed locally, the proposed algorithm uses the symmetrical properties of indoor artificial objects to estimate the occluded parts to obtain more accurate quadric parameters. Experiments have shown that compared with the state-of-art algorithm, especially on the forward trajectory of mobile robots, the proposed algorithm significantly improves the accuracy and convergence speed of quadric reconstruction. Finally, we made available an opensource implementation to replicate the experiments.

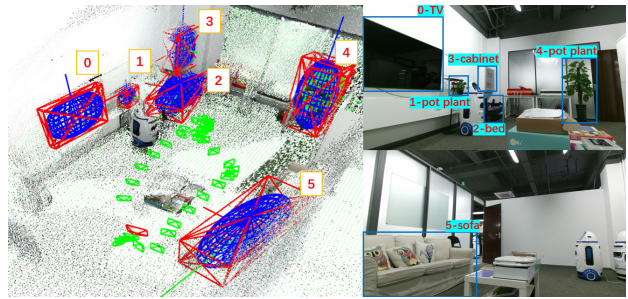

Figure 1. Object reconstruction on a mobile robot. *The image in the left shows the reconstruction result of 6 objects represented by the quadric model built by fusing object detection and RGB-D data. The objects include TV, pot plant, bed, cabinet, and sofa as indoor common objects. The blue ellipsoids visualize the estimated quadric model, and the red lines show their circumscribed cuboids for better visualization of their shape and orientation. The green rectangles show the trajectory of the robot. See the Experiment part for detail.*

## I. INTRODUCTION

Visual Simultaneous Localization and Mapping (SLAM) estimates the camera state and builds the environment map at the same time, which has gained rapid development and wide application in the past several decades [1]. However, the landmarks of traditional visual SLAM algorithms are usually based on points, lines, planes and other low-dimensional geometric features, which seldom involve semantic information and man-made structural regularities in indoor scenarios. Considering the working environment of an indoor mobile robot, the environmental structure is usually characterized by regular geometry (various furniture and electrical appliances) distributed in the closed space (wall, ground, and ceiling). Using this feature to realize the indoor SLAM algorithm of a robot, we could simplify the map representation, and also improve its robustness to the environment change (such as the environmental light change). In addition, if the map contains the semantic and volume information of objects, it will be conducive to the realization of semantic navigation, object grasping and other advanced tasks of mobile robots.

With the rise of deep learning in recent years, researchers began to introduce object landmarks into the SLAM algorithm. For example, Maskfusion [2] is an object-level SLAM system using Mask-RCNN [23] as the front-end to realize the instance-level object detection and segmentation, and the Surfels model to store three-dimensional objects. Similarly, Fusion++ [3] uses the TSDF model, and MID-Fusion [4] uses the Octree-Voxel model to achieve the same function. However, those SLAM algorithms use dense object models, trying to reconstruct the fine texture information of the object surface, which leads to very high demand for computing and storage resources. The reconstruction of fine object surface may be more suitable for virtual reality applications. For most applications of indoor robots, such as robot navigation, grasping, etc., it is not necessary.

In this regard, QuadricSLAM [5] proposed for the first time to introduce quadric representation into SLAM as a mathematical description of objects, and only relied on the bounding boxes generated by object detection algorithms to realize the estimation of quadric parameters. As a sparse object model, the quadric surface, such as ellipsoid, models the object position, orientation and occupied space. And the projection of the dual quadric surface has detailed mathematical theory support [14]. The following research also proves that the quadric surface can model the support relationship with spatial structure compactly [7], and it can introduce prior information such as object size [10], orientation [11] conveniently, which is an ideal mathematical representation for modeling environmental objects.

QuadricSLAM is based on a monocular camera, and the initialization of the quadric surface needs to fuse the object detection results of images from at least three perspectives.

The research was supported by the National Natural Science Foundation of China (No. 91748101).

[1]Robotics Institute, Beihang University (BUAA), Beijing, China. (E-mail: {liaoziwei, wangweilab, qixianyu, zhang_xy}@buaa.edu.cn).

[2]Beijing Evolver Robotics Technology Co., Ltd, Beijing, China. ( E-mail : { xuelin, jiaojianzhen, weiran}@ren001.com ).

*Correspondence author.

However, it is difficult to generate a variety of viewing angles under the plane trajectory of the mobile robot, making the initialization of the quadric surface difficult. Especially, lacking the change of the pitch angle of the object makes it is easy to generate the unobservable problem. In order to solve this problem, Ok et al. [10] proposed to further constrain the ellipsoid by estimating the texture plane on the surface of the object, but this method needs the prior information of the size of the object and requires that the surface of the object has obvious texture.

In the previous work, the authors have studied the extraction of indoor environmental planar features and its application in visual SLAM [13]. This paper further focuses on the characteristics of indoor objects and proposes a sparse object-level SLAM algorithm using the quadric surface as the object model. In order to solve the observability problem under the plane trajectory of the mobile robot, this paper proposes to fuse data from commercial RGB-D camera commonly used by mobile robots. By integrating object detection results and depth data, the proposed algorithm uses single frame data to realize the initialization of quadric surface and adds it as an additional three-dimensional constraint into the graph optimization. Aiming at the problem of point cloud missing when the object is partially occluded by itself, the algorithm estimates the symmetry plane of the object based on the symmetry assumption of most indoor artificial objects. Then it completes the point cloud to obtain more accurate quadric surface parameters. We use the public TUM-RGB-D dataset and an indoor dataset of a real mobile robot to evaluate the effect of quadric reconstruction and prove the effectiveness of the above methods. The code is public available[1].

## II. RELATED WORK

### A. Landmark representations in SLAM

Traditional visual SLAM often uses point features [16-19]. In recent years, SLAM algorithms based on line features [21] and plane features [13] [15] [20] have been proposed to enhance the system robustness and further improve the accuracy. However, these slam algorithms based on the low dimensional geometric model lack semantic information in the map, which is not conducive to the semantic navigation and high-level human-computer interaction with the mobile robot and limits the scene understanding ability of the mobile robot.

### B. Quadric object models in SLAM

Quadric surface model [14], such as ellipsoid, as a sparse object representation model, has a complete mathematical theory and compact form, which can model the position, occupying space and orientation of objects. Existing work [29] [30] [31] explored the reconstruction of quadric surface based on the bounding boxes of object detection, but it was limited to landmark reconstruction and did not estimate camera pose. QuadricSLAM [5] is the first attempt to introduce quadric representation as an object landmark into the graph optimization of SLAM, and build the object-level environment map while estimating the pose. Then many works are proposed on this basis: Jablonsky et al. [11] proposed the use of gravity prior to constrain ellipsoid; Ok et al. [10] explored the object size prior; Hosseinzadeh et al. [7] realized a SLAM algorithm of point, plane, and ellipsoid, and put forward the constraint relationship between the supporting plane and ellipsoid; Gaudilliere et al. [28] introduced the re-localization based on quadric surface.

Besides, it is worth mentioning that Yang et al. proposed CubeSLAM [6] to estimate the objects with the cube model. The cube model is recovered from the single monocular image by using the bounding boxes generated by the object detection algorithm, and the vanish points. However, the calculation of vanishing points needs the extraction of parallel lines from the object surface, which limits the applicable object categories.

### C. Object model estimation from RGB-D data

In the field of robot's object grasping, the quadric surface model has been widely used in the analysis of grasping points. There are many research results of estimating the quadric surface based on RGB-D camera data. For example, a method of fitting quadric surface from a single RGB-D frame is proposed in [25], and a priori information of object size is introduced into the fitting process in [27]. Considering the common problem of point cloud missing when the object is partially occluded by itself, Schiebener et al. [24] proposed to integrate the supporting plane and symmetry of the object to complete the point cloud. Makhal et al. [26], based on the point cloud after the completion of symmetric surface estimation, obtain the quadric surface model by fitting for object grasping. The above work generally assumes that the point cloud segmentation has been completed, and the algorithm proposed in this paper will combine the object detection and supporting plane estimation to complete the point cloud segmentation. Also, the above work only uses single frame RGB-D data, and in this paper, the ellipsoid estimation results are integrated into the graph optimization framework of SLAM to achieve multiple frames fusion.

## III. MATHEMATICS OF QUADRIC MODEL

There are many kinds of quadric surfaces: ellipsoid, hyperboloid, cylinder, etc. Considering the closed geometry of indoor artificial objects, the ellipsoid in quadric surface is used as the object model. In the existing work [29] [30] [31], the parametric model and projection geometry of the quadric surface are discussed in detail, and this part gives a brief overview.

### A. Quadric surface and its projection

The quadric surface in 3D space can be described by 4x4 symmetric matrix $Q$. Any 3D point $X$ on the surface of the quadric meets the condition:

$$X^T Q X = 0. \quad (1)$$

To model the projection process conveniently, the dual form of the quadric surface is introduced. For point quadric surface $Q$, there is a dual expression form $Q^*$, for any plane $\pi$ tangent to the quadric surface, it satisfies the following equation: $\pi^T Q^* \pi = 0$. The matrix $Q^*$ is the adjoint matrix of $Q$. If $Q$ is

---
[1] https://github.com/XunshanMan/Object-oriented-SLAM

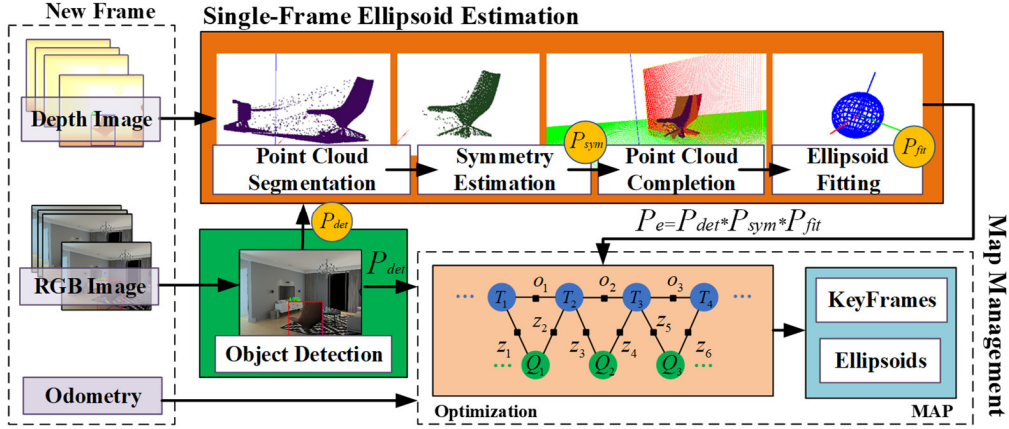

Figure 2. System overview. The *RGB image, depth image of an RGB-D camera and odometry data are system input. The depth image and the object detection result are fused to complete the single-frame ellipsoid estimation. Object detection result, estimated ellipsoids and odometry will be put into the back-end optimization together. In Optimization, the nodes are composed of robot state T and object Q expressed by quadric representation. The edges are composed of camera observation Z and odometry constraint O. More detail can be seen in part IV and V. The picture material is from [36].*

invertible, it can be expressed as:

$$Q^* = |Q|Q^{-1}. \quad (2)$$

The dual quadric surface $Q^*$ is projected under the camera observation $P$ to obtain the curve $C^*$. This process can be compactly expressed as:

$$C^* = PQ^*P^T, \quad (3)$$

where $P = K[R|t]$, $K$ is the internal parameter of the camera, and $R$ and $t$ are camera rotation and translation. In particular, when the quadric surface is an ellipsoid, the projection curve $C^*$ is an ellipse. In the process of parameter optimization of ellipsoid described later, two-dimensional matching constraints will be constructed between the minimum enclosing rectangle of $C^*$ and the rectangle obtained from object detection.

*B. Decomposition of quadric surface*

In order to obtain the position and size of the ellipsoid, the dual form parameter matrix $Q^*$ of the ellipsoid can be decomposed by eigen-decomposition:

$$Q^* = Z\breve{Q}^*Z^\top. \quad (4)$$

In the result on the right, $\breve{Q}^*$ is the dual quadric surface with coordinates at the origin and no rotation; $Z \in SE(3)$ is a homogeneous transformation, including the spatial position and orientation of quadric surface.

$$Z = \begin{pmatrix} R(\theta) & t \\ 0_3^T & 1 \end{pmatrix} \text{ and } \breve{Q}^* = \begin{pmatrix} s_1^2 & 0 & 0 & 0 \\ 0 & s_1^2 & 0 & 0 \\ 0 & 0 & s_1^2 & 0 \\ 0 & 0 & 0 & -1 \end{pmatrix} \quad (5)$$

Among them, $s_1, s_2, s_3$ are the size of the semi-axis length of the ellipsoid principal axis. An ellipsoid in space contains nine free parameters: translation, rotation (described by Euler angles), and scale, which can express the position, orientation and three-dimensional occupied space of the object. This shows that the dual form can be used to express the ellipsoid conveniently, which is conducive to subsequent calculation and visualization.

IV. SINGLE-FRAME ELLIPSOID ESTIMATION BASED ON AN RGB-D CAMERA

This section will introduce the process of ellipsoid parameter estimation based on a single frame observation of an RGB-D camera. The whole system can be divided into several steps, such as point cloud segmentation, symmetry plane estimation, point cloud completion and ellipsoid fitting, as shown in Figure 2.

*A. Point cloud segmentation*

Considering that the objects in the indoor environment are generally supported on a large plane (such as the ground, desktop, etc.) and perpendicular to the supporting plane. The proposed algorithm combines object detection and the supporting plane to segment the object point cloud and gets the outer envelope point cloud. The initial camera orientation relative to the ground is assumed to be known, which is usually parallel to the ground for most mobile robots, so the rough gravity direction can be attained to help extract the supporting plane. The point cloud segmentation process of single-frame RGB-D data is as follows.

First, use RANSAC to extract potential supporting plane in depth image based on the author's previous work on plane extraction [13]. Two criteria are set to filter invalid planes: angles between the normal vector and the gravity direction must be less than $\varepsilon_0$; the number of points in the plane must be larger than $\varepsilon_1$. The remaining planes are marked in a set $S = \{\pi_k\}$;

Then, detect a group of bounding boxes $\{b_i\}$ from RGB images using the object detection algorithm (such as Yolo [34]), and execute 1) to 3) for each bounding box $\{b_i\}$:

1) Extract the point cloud $C_i^a$ in the bounding box region $b_i$ from the depth image. Because of the viewing angle, $C_i^a$ generally includes some points that belong to the supporting plane;

2) Select the plane from $S$ whose distance to the point cloud center is minimum as the supporting plane. To make sure the supporting plane is below the object, this paper also considers the direction of the distance. Let $d(P, \pi)$ represent the distance between point $P$ and plane $\pi$. Assumed that positive symbol means the point lies above, then the supporting plane $\pi_i^{sup}$ must meet the condition:

$$\pi_i^{sup} = \min_{\pi} d(P_i, \pi), \pi \in \{\pi \mid d(P_i, \pi) > 0, \pi \in S\}$$

3) Filter out those points that are close to or below the supporting plane $\pi_i^{sup}$ and obtain the point cloud $C_i^b$ that lies above the plane.

$$C_i^b = \{P \mid d(P, \pi_i^{sup}) < \varepsilon_2, P \in C_i^a\}$$

Sometimes, there are still some outliers, e.g. the TV cabinet behind the chair in Figure 1.

4) Apply Euclidean clustering for $C_i^b$ and get a group of clusters. We choose the cluster that has minimum distance to the point generated from the center of the bounding box $b_i$ in the depth image. Choose it as the corresponding envelope point cloud $C_i$ if it has a larger number of points than threshold $\varepsilon_3$.

So far, for a given frame of RGB-D data, the bounding box of each object in the image and the corresponding envelope point cloud $\{b_i, C_i\}$ are obtained. Three threshold values are needed according to the resolution of the RGB-D camera. In the experiments, we set $\varepsilon_0$ =10deg, $\varepsilon_1$ =200, $\varepsilon_2$ =0.05m, $\varepsilon_3$=100 for Kinect2 camera.

*B. Point cloud completion using symmetry*

Due to occlusion and limited viewing angle, the point cloud $C_i$ only contains a part of the object. To obtain more complete object point cloud and improve the accuracy of quadric surface fitting, this paper introduces the general symmetry property of artificial objects in the indoor environment.

Based on the assumption of the ellipsoid model, there are two types of object symmetry: plane reflection and dual plane reflection, as shown in Figure 3. An algorithm for symmetry plane estimation has been proposed in previous work [9]. Firstly, samples are used to traverse the symmetry type and the corresponding symmetry searching domain, and then the local optimization is done to solve the fine symmetry parameters. For a given object point cloud $C$, a probability $P_{sym}$ is defined to every sampled symmetry plane $\pi$ to describe the symmetry score in [9]. This paper follows the probabilistic model, and improved the traversal process to accelerate the operation speed and meet the real-time requirements of a SLAM algorithm.

As shown in Table 1, the symmetrical forms of artificial objects such as TV, chairs, and laptops are already known and remain universal. By using the semantic labels of object detection, the searching domain of symmetry parameters is directly determined by finding the symmetry prior table.

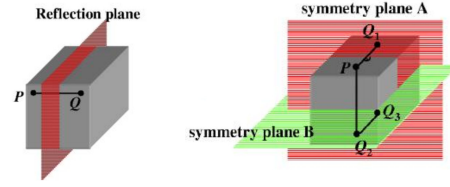

Figure 3. Plane reflection and dual plane reflection [9].

TABLE 1 SYMMETRY CATEGORIES OF COMMON OBJECTS IN INDOOR SCENES

| Symmetry Types | Searching Domains | Labels |
|---|---|---|
| Plane Reflection | $\pi_1$ | Chair |
| | | Laptop |
| | | Cup |
| Dual Plane Reflection | $\pi_1, \pi_2$ ($\pi_1 \perp \pi_2$) | Table |
| | | TV |
| | | Keyboards |

To quickly determine the initial symmetry plane, we propose to use the ellipsoid model assumption instead of performing equal spacing sampling in [9]: a) since the object is perpendicular to the support plane, according to the ellipsoid object model, the symmetry plane of the object must be perpendicular to the support plane; b) we use the principal component analysis (PCA) to point cloud $C_i$ and obtain the main direction of the object in the support plane. Then we take it as the normal direction of the initial symmetry plane. c) we assume that the symmetry plane passes through the center of the point cloud $C_i$. In this way, the initial symmetry plane can be fully determined and then is sent to local fine optimization.

After getting the symmetry plane $\pi_i^{sym}$, all points in the point cloud are mirrored according to the plane to get the mirrored point cloud. The original point cloud and the mirrored point cloud together constitute the complement point cloud $\breve{C}_i$, which will participate in the ellipsoid fitting process. The symmetry probability $P_{sym}$ corresponding to the symmetry plane $\pi_i^{sym}$ will participate in the final ellipsoid probability calculation.

*C. Ellipsoid fitting*

After the completion of point cloud $\breve{C}_i$, the parameter $Q_i^c$ of the ellipsoid under the observation of this frame is obtained by using the least square method. For any point $X \in \breve{C}_i$, the distance from the point x to the surface of ellipsoid $Q_i^c$ can be described by the following formula:

$$F(X, Q_i^c) = X^T Q_i^c X \quad (6)$$

To estimate an ellipsoid surface $Q_i^c$ containing all surface points as much as possible, the following least square problem can be defined:

$$Q_i^c = \min_Q \sum_{i=1}^N \left(\sqrt{s_1 s_2 s_3}(F(X_i, Q))\right)^2 \quad (7)$$

where $s_1 s_2 s_3$ is the volume of the quadric surface, which is obtained by decomposition from $Q_i^c$. The purpose of using volume as an adjustment parameter is to limit the volume of ellipsoid not to be too large. In practice, the least square problem is solved by the Levenberg Marquardt method or the Gauss-Newton method. We calculate a fitting probability $P_{fit}$ for the estimated ellipsoid defined by the residual:

$$P_{fit} = |2\pi|^{-\frac{1}{2}} \exp -\frac{1}{2} T \quad (8)$$

where the residual $T = \sum_{i=1}^{N} \frac{(\sqrt{s_1 s_2 s_3}(F(X_i, Q_i^c)))^2}{N}$.

Using the obtained point form parameters of the ellipsoid, the dual form of $Q_i^{c*}$ can be calculated by Equation (2). So far, the algorithm has obtained a set of objects data $\{b_i, Q_i^{c*}\}$ from the incoming single-frame RGB-D data. Each object data contains the object detection bounding box $b_i$ and the corresponding dual ellipsoid parameter $Q_i^{c*}$, which will participate in the graph optimization of SLAM as observation constraints. The detection probability $P_{det}$ is given by the object detection algorithm. The probability $P_e$ of the ellipsoid $Q_i^{c*}$ is obtained by the combination of the detection probability $P_{det}$, the symmetry estimation probability $P_{sym}$ and the ellipsoid fitting probability $P_{fit}$, as follows:

$$P_e = P_{det} P_{sym} P_{fit} \quad (9)$$

The probability will influence the weights of those observations in the following optimization process.

V. SLAM BASED ON THE QUADRIC SURFACE OBJECT MODEL

A. Problem definition

The complete object-level SLAM algorithm can be represented by the factor graph shown in Figure 2. Therefore, the SLAM problem expressed in this graph can be modeled as the following nonlinear optimization problem:

$$\hat{T}, \hat{Q} = \arg\min_{T,Q}(\sum H_z(f_z) + \sum H_o(f_o)) \quad (10)$$

Huber kernel is chosen to enhance the robustness towards outlier observations in $H(\cdot)$. $f_o$ is the odometry constraint. $f_z$ is the camera-object observation constraint that consists of two parts: the three-dimensional constraint $H_{z_3}(f_{z_3})$ based on the single-frame ellipsoid estimation and the two-dimensional constraint $H_{z_2}(f_{z_2})$ based on object detection, as follows:

$$H_z(f_z) = \varepsilon_z H_{z_3}(f_{z_3}) + H_{z_2}(f_{z_2}) \quad (11)$$

The parameter $\varepsilon_z$ is used to adjust the weight of two-dimensional and three-dimensional observation constraints. The experiments will give an evaluation of the effect of the parameter.

B. 3D object observation constraints

For one incoming frame of RGB-D data, through the single-frame ellipsoid estimation algorithm in IV, the dual form of the estimated ellipsoid in the local coordinate system can be obtained. The dual form can easily express the camera observation process. The expression of $Q_i^{w*}$ in the world coordinate system can be obtained by transforming $Q_i^{c*}$:

$$Q_i^{w*} = H_{wc} Q_i^{c*} H_{wc}^T \quad (12)$$

where $H_{wc}$ is the transformation matrix from camera coordinate to world coordinate corresponding to the current frame position $x_i$. If the observed object has been initialized in the map, the corresponding ellipsoid $Q_i^{m*}$ of ellipsoid $Q_i^{w*}$ can be found in the map. The process of searching is a data association problem. Some existing work [33] proposed semantic data association algorithms to the problem. This paper will not go deep into it and assume the data association is solved, to focus on the performance of single-frame ellipsoid estimation. Therefore, the three-dimensional object observation constraint is the constraint between $Q_i^{w*}$ and $Q_i^{m*}$, which is defined as follows:

$$f_{z_3}(Q_i^{w*}, Q_i^{m*}) = P_e \|v(Q_i^{w*}) - v(Q_i^{m*})\|_\Sigma^2 \quad (13)$$

Among them, $P_e$ represents the probability given by the single-frame ellipsoid estimation process, which is obtained by integrating object detection probability, symmetry probability, and ellipsoid fitting probability. $v(Q^*)$ represents the vector expression of the dual form ellipsoid $Q^*$, i.e. translation, rotation (Euler angle description), scales, obtained from 4x4 matrix

$$v(Q^*) = [x, y, z, roll, pitch, yaw, s_1, s_2, s_3]^T \quad (14)$$

C. 2D object observation constraints

The two-dimensional object observation constraint is the constraint between the object detection bounding boxes and the projection of ellipsoid on the image plane, which is constructed in the pixel plane of the image. It is complementary with three-dimensional observation constraints to improve the robustness of the system.

The two-dimensional object observation constraint in the system follows the description in QuadricSLAM [5], where it is considered that the four edges of the object detection rectangle are tangent to the ellipse projected by the ellipsoid onto the image plane. The projection matrix $P_j$ of the camera can be obtained by knowing the current pose $x_j$. By the Equation (3), the ellipsoid in the map can be projected into the image plane of the current frame position $x_j$ to obtain the elliptic curve $C^*$, and then the circumscribed rectangle $\beta(C^*)$ can be generated according to the elliptic curve $C^*$. Considering a given object detection bounding box $b_i$, and the corresponding ellipsoid in the map, the constraint functions of two-dimensional objects can be defined as follows:

$$f_{z_2}(x_j, b_i, Q_i^{m*}) = P_{det} \|b_i - \beta(C^*)\|_\Sigma^2 \quad (15)$$

Among them, $P_{det}$ is the object detection probability given by the object detection algorithm, and both the rectangle is described by the pixel coordinates of the upper left vertex and the lower right vertex.

D. Odometry constraints

The input of odometry constraint can be the wheel odometry of a mobile robot, or the output of visual odometry algorithm such as ORB-SLAM [16]. Define $d(x, u)$ to calculate the result of pose x after motion u, then the odometry constraint can be formulated as:

$$f_o(x_j, u_j, x_{j+1}) = \|d(x_j, u_j) \ominus x_{j+1}\|_{\Sigma_j}^2 \quad (16)$$

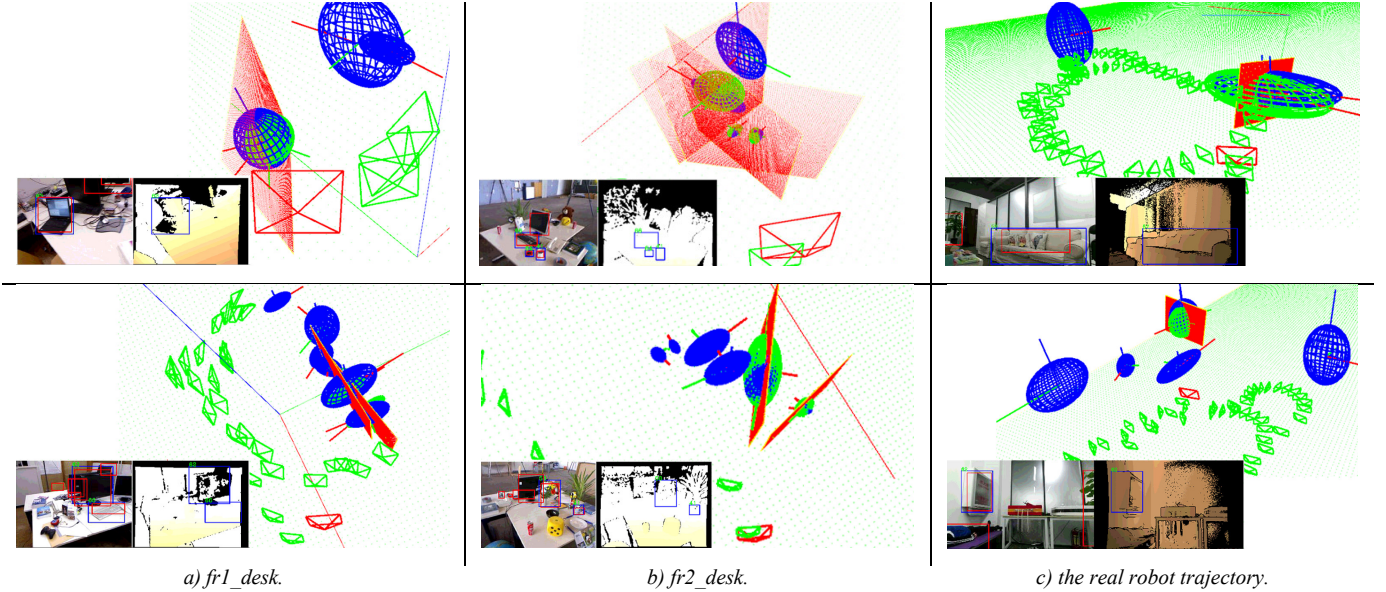

*a) fr1_desk.*     *b) fr2_desk.*     *c) the real robot trajectory.*

Figure 4. The reconstruction process of DwB on the TUM-RGBD dataset and the real robot trajectory. *The green rectangles show the camera trajectory, and the red one shows the current camera pose. The blue ellipsoids show the objects in the map. The green ellipsoids show the result of the single-frame ellipsoid estimation, and the red planes show the result of the symmetry plane estimation, both in the current frame. On RGB images, blue rectangles show the YOLO object detection result, and red ones show the projection of the ellipsoids in the map. It is recommended to read in the color version.*

where $\Sigma_j$ is the covariance matrix of odometry in this frame.

## VI. EXPERIMENTS

The experiment focused on the improvement of ellipsoid mapping accuracy and convergence speed after the introduction of three-dimensional observation constraints. The results are compared under the orbit and forward motion modes respectively. The ground-truth trajectory is used as the input to maximize the mapping effect.

Experiments for orbit motion mode used the public TUM-RGB-D dataset, where a Kinect2 RGB-D camera was held in hand and moved around a desktop with objects lied on for one circle, which generated observations with large angle of view changes. The ground-truth trajectory is already given by the dataset. Experiments under the forward motion mode used a real mobile robot trajectory, where a turtlebot3 traveled in an indoor home-like environment with a Kinect2. The ground-truth trajectory was obtained from the fusion of ORB-SLAM2 and wheel odometry data with several loop closures. To get the ground-truth object parameters, we used ElasticFusion [35] to build a high-precision dense point cloud, then manually annotated the poses and scales of the real objects.

QuadricSLAM is selected as the baseline, which is a state-of-art object-level SLAM system using quadrics as the object landmark and is based on a monocular camera. Because the proposed algorithm introduces the supporting plane, to ensure the fairness, we use the related version of QuadricSLAM in [7], which also introduces the supporting plane constraint, to directly compare the effectiveness of the three-dimensional constraints introduced by the single-frame ellipsoid estimation. We used Yolo [34] as the object detection algorithm in the experiments. We manually annotated data association and filtered the bounding boxes lying on the image edges to find the limitation of our system performance.

TABLE 2 THE ELLIPSOID ESTIMATION RESULTS

| Sequences | Items | QuadricSLAM | DO | DwB |
|---|---|---|---|---|
| Fr1_desk (3)[a] | Shape | 0.399557 | **0.27516** | 0.347189 |
| | Trans (m) | 0.056574 | 0.12403 | **0.05305** |
| | Rot (deg) | 26.2496 | 5.84448 | **4.64115** |
| Fr1_desk2 (4) | Shape | 0.549205 | **0.46343** | 0.479532 |
| | Trans | 0.078322 | 0.199142 | **0.07743** |
| | Rot (deg) | 16.6234 | 12.1614 | **10.5497** |
| Fr2_desk (5) | Shape | 0.423198 | 0.384096 | **0.26592** |
| | Trans (m) | **0.08219** | 0.106938 | 0.08449 |
| | Rot (deg) | **10.6924** | 15.3503 | 13.8469 |
| TUM Average (12) | Shape | 0.45929 | 0.383305 | **0.35744** |
| | Trans (m) | 0.074495 | 0.141946 | **0.07428** |
| | Rot (deg) | 16.5587 | 11.91088 | **10.4464** |
| Real_robot (6) | Shape | 0.609898 | 0.41569 | **0.2766** |
| | Trans(m) | 0.54414 | 0.18893 | **0.18451** |
| | Rot(deg) | 21.3404 | **18.9657** | 19.5091 |

a. The number of valid objects in the trajectory (Objects containing at least 5 keyframe observations with a probability of over 0.95.).

To fully evaluate the position, orientation and shape of the reconstructed objects, the evaluation benchmark in the experiments used Trans (m) to measure the average distance between the estimated object and the real object, and Rot (deg) to measure the average angular offsets of the main axis of the objects after projecting them on the supporting plane. The evaluation for shape used the Jaccard distance (1 - Intersection over Union) between the 3D axis-aligned bounding box of the estimated ellipsoid and the real bounding box after they are translated to the origin point.

## A. The TUM-RGB-D dataset

This part of the experiments cover three trajectories from the TUM-RGB-D dataset: fr1_desk, fr1_desk2, and fr2_desk. To measure the accuracy limit, we selected the object detection bounding boxes from YOLO with a probability of more than 0.95 and selected the objects that could be effectively observed for at least 5 times as the valid objects. See Table 2 for the number of valid objects in each trajectory.

We designed two types of algorithms for evaluation: The Depth with Bounding-boxes (DwB) algorithm, which combined both the three-dimensional constraints and the two-dimensional constraints; and The Depth Only (DO) algorithm, which activated the three-dimensional constraints only. The baseline QuadricSLAM only contained two-dimensional constraints.

The results in Table 2 show that DwB achieved a better shape and rotation accuracy than QuadricSLAM in the experiments. Especially in trajectory fr1_desk and fr1_desk2, the shape and rotation accuracy was significantly improved, because the two trajectories contained large symmetrical objects such as computer monitors, keyboards and laptops, as shown in Figure 4.a. The successful estimation of the symmetry plane helped improve the rotation accuracy.

Although the shape of the object was improved in the f2_desk dataset, the rotation shows negative growth, because the trajectory contained smaller objects such as a mouse, teacup that are difficult to estimate the symmetry, and also contained objects without symmetry such as pot plants, as in Figure 4.b.

Considering the translation, it stays almost unchanged, and DO even shows negative growth, which demonstrates that the two-dimensional constraint with large perspective angle changes could achieve an accurate estimation of the object center. One single-frame RGB-D data can only see a part of the object, although there is a symmetry complement, it is still difficult to accurately estimate the center position of the object. DwB, which considered both 3D and 2D constraints, added their advantages together and successfully improved rotation and shape accuracy while maintaining translation accuracy.

## B. A real mobile robot trajectory in an indoor scene

Mobile robots are difficult to generate orbit trajectories. They move forward most of the time, with limited changes of view angles towards objects. This experiment covered a trajectory recorded by a turtlebot3 robot mounting a Kinect2 in an indoor environment. The environment contained a TV, sofa, bed and other common indoor furniture, as shown in Figure 1. In the trajectory, the observation angle of the object changed little, and the number of effective observations was very small, which was a huge challenge for QuadricSLAM as it only uses object detection as the information resource.

From the results of real robot experiment in Table 2, we can see that under the forward trajectory, QuadricSLAM suffered a big loss on accuracy. The average shape error is 0.61 and the translation error is 0.54 m. Due to the limited observation angle, the estimated ellipsoid is easy to be unobservable on some axes, leading to extreme big or small ellipsoid estimation, as shown in Figure 5.1.

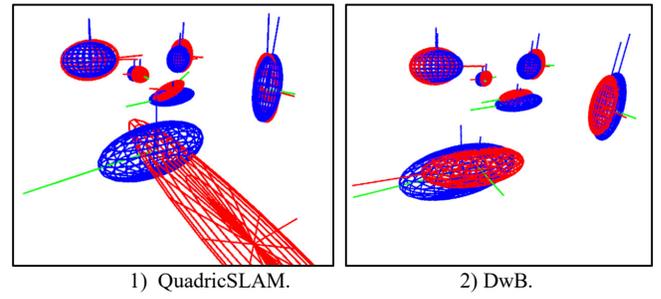

1) QuadricSLAM.  2) DwB.

Figure 5. Results compared with ground-truth on a forward trajectory. *Red shows estimated ellipsoids, and blue shows ground-truth ellipsoids.*

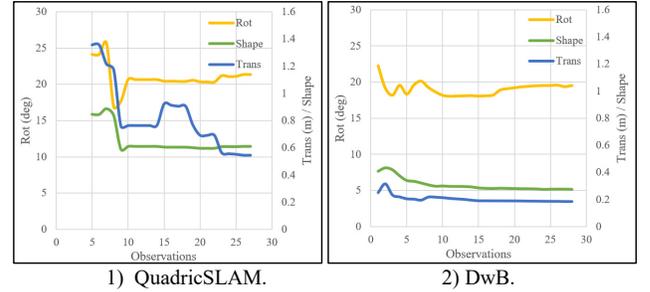

1) QuadricSLAM.  2) DwB.

Figure 6. With the increase of observation times, the curve tends to converge. *QuadricSLAM becomes stable after about 25 frames. The initial error of DwB is already small, and then it converges slowly after 10 frames. The final result of DwB is also better.*

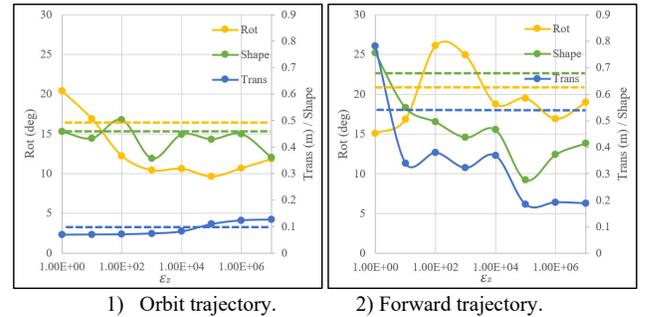

1) Orbit trajectory.  2) Forward trajectory.

Figure 7. Analysis of scale parameter $\varepsilon_z$. *Scale parameter $\varepsilon_z$ adjusts the proportion between 2D and 3D constraints in the optimization process. In the orbit trajectory, 2D observation is complete and accurate, and extremum tends to be obtained at a lower value of $\varepsilon_z$, giving 2D constraints more proportion; in the forward case, 2D constraint only provides partial axis constraint for an ellipsoid, and extremum is obtained at a higher value of $\varepsilon_z$, giving 3D constraints more proportion. The dash lines represent the result of QuadricSLAM.*

The performance of the single-frame ellipsoid estimation was still stable even under a small viewing angle. Compared with QuadricSLAM, the shape and translation accuracy was significantly improved, especially for DwB. Considering the rotation, large objects such as TV, sofa, bed, and cabinet that was successfully estimated the symmetry, achieved a high accuracy of rotation angle estimation, as shown in Figure 5.2. However, due to the confusion of the symmetry estimation of the small pot plant, the average accuracy of the rotation angles remain low.

In addition, we further analyzed the convergence rate of QuadricSLAM and DwB after the introduction of 3D constraints. The convergence curves were drawn with the observation times of the objects as the horizontal axis and the average Trans, Rot, Shape errors as the vertical axis, as shown in Figure 6. As QuadricSLAM need at least three observations to initialize a new ellipsoid, ours was able to finish the initialization with only one observation relatively precisely. There was a violent oscillation in the early stage of QuadricSLAM. It gradually converged and became stable after about 25 frames of observations. Comparatively, as the initial estimation was accurate enough, DwB converged and became stable after about 10 frames, and the error difference between the convergence value and the initial value was small, which reflected the performance of rapid and accurate initialization of the algorithm.

*C. Parameter analysis*

When combining 3D and 2D constraints together into the optimization process, weight param $\varepsilon_z$ influences the proportion of 3D and 2D constraints. The relationship between the choice of param $\varepsilon_z$ and the average error of all objects after convergence was drawn in Figure 7. The result shows that better performance can be obtained under around $10^3$ under orbit trajectory. And it shows that under forward trajectory, better performance can be obtained under around $10^5$. The result shows that in the orbit trajectory where 2D constraints are relatively fully constrained, the proportion of 2D constraints can be strengthened, otherwise, in the forward trajectory where 2D constraints involve observability problem, it will be better to give 3D constraints more weight.

## VII. CONCLUSION

This paper proposes a sparse object-level SLAM algorithm based on an RGB-D camera for mobile robots in an indoor environment. By using the quadric representation as the object model, the position, orientation and occupied space of the object are expressed compactly. We propose an effective method to solve the observability problem under the planar motion trajectory of a mobile robot, by fusing the depth data with object detection to estimate the ellipsoid parameters.

Considering the future work, more constraints can be explored at the object level, such as the semantic relationship between objects and spatial structures; the introduction of semantic data association [33] can make the system more robust; as the external cuboid of an ellipsoid can represent the model proposed in CubeSLAM [6], it will be interesting to explore the two system together in the future.

REFERENCES

[1] Cadena, C., Carlone, L., Carrillo, H., Latif, Y., Scaramuzza, D., Neira, J., … Leonard, J. J. (2016). Past, present, and future of simultaneous localization and mapping: Toward the robust-perception age. IEEE Transactions on Robotics, 32(6), 1309–1332.
[2] Runz, M., Buffier, M., & Agapito, L. (2019). MaskFusion: Real-Time Recognition, Tracking and Reconstruction of Multiple Moving Objects. Proceedings of the 2018 IEEE International Symposium on Mixed and Augmented Reality, ISMAR 2018, 10–20.
[3] McCormac, J., Clark, R., Bloesch, M., Davison, A., & Leutenegger, S. (2018). Fusion++: Volumetric object-level SLAM. In Proceedings - 2018 International Conference on 3D Vision, 3DV 2018 (pp. 32–41).
[4] Xu, B., Li, W., Tzoumanikas, D., Bloesch, M., Davison, A., & Leutenegger, S. (2018). MID-Fusion: Octree-based Object-Level Multi-Instance Dynamic SLAM. Retrieved from http://arxiv.org/abs/1812.07976
[5] Nicholson, L., Milford, M., & Sunderhauf, N. (2019). QuadricSLAM: Dual quadrics from object detections as landmarks in object-oriented SLAM. IEEE Robotics and Automation Letters, 4(1), 1–8.
[6] Yang, S., & Scherer, S. (2019). CubeSLAM: Monocular 3-D Object SLAM. IEEE Transactions on Robotics, 35(4), 925–938.
[7] Hosseinzadeh, M., Latif, Y., Pham, T., Suenderhauf, N., & Reid, I. (2018, December). Structure Aware SLAM Using Quadrics and Planes. In Asian Conference on Computer Vision (pp. 410-426). Springer, Cham.
[8] Hosseinzadeh, M., Li, K., Latif, Y., & Reid, I. (2019, May). Real-time monocular object-model aware sparse slam. In 2019 International Conference on Robotics and Automation (ICRA) (pp. 7123-7129). IEEE.
[9] Thrun, S., & Wegbreit, B. (2005, October). Shape from symmetry. In Tenth IEEE International Conference on Computer Vision (ICCV'05) Volume 1 (Vol. 2, pp. 1824-1831). IEEE.
[10] Ok, K., Liu, K., Frey, K., How, J. P., & Roy, N. (2019). Robust Object-based SLAM for High-speed Autonomous Navigation. Proceedings of the IEEE International Conference on Robotics and Automation (ICRA), 669–675.
[11] Jablonsky, N., Milford, M., & Sünderhauf, N. (2018). An Orientation Factor for Object-Oriented SLAM. Retrieved from http://arxiv.org/abs/1809.06977
[12] Redmon, J., Divvala, S., Girshick, R., & Farhadi, A. (2016). You only look once: Unified, real-time object detection. In Proceedings of the IEEE conference on computer vision and pattern recognition (pp. 779-788).
[13] Zhang, X., Wang, W., Qi, X., Liao, Z., & Wei, R. (2019). Point-Plane SLAM Using Supposed Planes for Indoor Environments. Sensors, 19(17), 3795.
[14] R. Hartley and A. Zisserman, Multiple View Geometry in Computer Vision. Cambridge, U.K.: Cambridge Univ. Press, 2003.
[15] Kaess, M. (2015). Simultaneous localization and mapping with infinite planes. In ICRA (Vol. 1, p. 2).
[16] Mur-Artal, R., Montiel, J. M. M., & Tardos, J. D. (2015). ORB-SLAM: A Versatile and Accurate Monocular SLAM System. IEEE Transactions on Robotics, 31(5), 1147–1163.
[17] Engel, J., Koltun, V., & Cremers, D. (2018). Direct Sparse Odometry. IEEE Transactions on Pattern Analysis and Machine Intelligence, 40(3), 611–625.
[18] J. Engel, J. Sch¨ops, and D. Cremers. LSD-SLAM: Large-scale direct monocular SLAM. In European Confonference on Computer Vision (ECCV), pages 834–849. Springer, 2014.
[19] Forster, C., Zhang, Z., Gassner, M., Werlberger, M., & Scaramuzza, D. (2017). SVO: Semidirect Visual Odometry for Monocular and Multicamera Systems. IEEE Transactions on Robotics, 33(2), 249–265.
[20] Yang, S., Song, Y., Kaess, M., & Scherer, S. (2016). Pop-up SLAM: Semantic monocular plane SLAM for low-texture environments. IEEE International Conference on Intelligent Robots and Systems, 2016–Novem, 1222–1229.
[21] Zhou, H., Zou, D., Pei, L., Ying, R., Liu, P., & Yu, W. (2015). StructSLAM: Visual SLAM With Building Structure Lines. IEEE Transactions on Vehicular Technology, 64(4), 1364–1375.
[22] Salas-Moreno, R. F., Newcombe, R. A., Strasdat, H., Kelly, P. H. J., & Davison, A. J. (2013). SLAM++: Simultaneous localisation and mapping at the level of objects. Proceedings of the IEEE Computer Society Conference on Computer Vision and Pattern Recognition, 1352–1359.
[23] He, K., Gkioxari, G., Dollár, P., & Girshick, R. (2017). Mask r-cnn. In Proceedings of the IEEE international conference on computer vision (pp. 2961-2969).
[24] Schiebener, D., Schmidt, A., Vahrenkamp, N., & Asfour, T. (2016). Heuristic 3D object shape completion based on symmetry and scene context. IEEE International Conference on Intelligent Robots and Systems, 2016–Novem, 74–81.
[25] Vezzani, G., Pattacini, U., & Natale, L. (2017). A grasping approach based on superquadric models. Proceedings - IEEE International Conference on Robotics and Automation, 1579–1586.
[26] Makhal, A., Thomas, F., & Gracia, A. P. (2018). Grasping unknown objects in clutter by superquadric representation. Proceedings - 2nd IEEE International Conference on Robotic Computing, IRC 2018, 2018–Janua, 292–299.
[27] Vezzani, G., Pattacini, U., Pasquale, G., & Natale, L. (2018). Improving Superquadric Modeling and Grasping with Prior on Object Shapes. Proceedings - IEEE International Conference on Robotics and Automation, 6875–6882.
[28] Gaudilliere, V., Simon, G., & Berger, M. O. (2019). Camera relocalization with ellipsoidal abstraction of objects. Proceedings - 2019 IEEE International Symposium on Mixed and Augmented Reality, ISMAR 2019, 8–18.
[29] Rubino, C., Crocco, M., & Del Bue, A. (2018). 3D Object Localisation from Multi-View Image Detections. IEEE Transactions on Pattern Analysis and Machine Intelligence, 40(6), 1281–1294.
[30] Crocco, M., Rubino, C., & Del Bue, A. (2016). Structure from Motion with Objects. Proceedings of the IEEE Computer Society Conference on Computer Vision and Pattern Recognition, 2016–Decem, 4141–4149.
[31] Gay, P., Bansal, V., Rubino, C., & Bue, A. Del. (2017). Probabilistic Structure from Motion with Objects (PSfMO). Proceedings of the IEEE International Conference on Computer Vision, 2017–Octob, 3094–3103.
[32] Vaskevicius, N., & Birk, A. (2019). Revisiting Superquadric Fitting: A Numerically Stable Formulation. IEEE Transactions on Pattern Analysis and Machine Intelligence, 41(1), 220–233.
[33] Bowman, S. L., Atanasov, N., Daniilidis, K., & Pappas, G. J. (2017). Probabilistic data association for semantic SLAM. Proceedings - IEEE International Conference on Robotics and Automation, 1722–1729.
[34] Redmon, J., Divvala, S., Girshick, R., & Farhadi, A. (2016). You only look once: Unified, real-time object detection. In Proceedings of the IEEE conference on computer vision and pattern recognition (pp. 779-788).
[35] Whelan, T., Leutenegger, S., Salas-Moreno, R. F., Glocker, B., & Davison, A. J. (2015). ElasticFusion: Dense SLAM without a pose graph. Robotics: Science and Systems, 11.
[36] W. Li, S. Saeedi, J. McCormac, R. Clark, D. Tzoumanikas, Q. Ye, Y. Huang, R. Tang, and S. Leutenegger, "Interiornet: Mega-scale multi- sensor photo-realistic indoor scenes dataset," in Proceedings of the British Machine Vision Conference (BMVC), 2018